\documentclass[runningheads]{llncs}
\usepackage{graphicx}

\usepackage{textcomp}
\usepackage{xcolor}
\usepackage{multirow}
\usepackage{subcaption}
\usepackage{cite}
\usepackage{amsmath,amssymb,amsfonts}
\usepackage{algorithmic}
\makeatletter
\newcommand{\smalloplus}{\mathbin{\mathpalette\make@small\oplus}}

\newcommand{\make@small}[2]{%
  \vcenter{\hbox{%
    \scalebox{1.0}{$\m@th#1#2$}%
  }}%
}
\makeatother

\begin{document}
\title{MPool: Motif-Based Graph Pooling}
%
%
\author{Muhammad Ifte Khairul Islam\inst{1} \and
Max Khanov\inst{2} \and Esra Akbas \inst{1}}
\authorrunning{Muhammad Ifte et al.}
%
\institute{Department of Computer Science, Georgia State University, GA,USA\\
 \and
 University of Wisconsin–Madison,Madison, Wisconsin, USA\\
\email{mislam29@student.gsu.edu,} \email{maximkha@outlook.com,}\email{eakbas1@gsu.edu}}

%
\maketitle              
\newcommand{\our}{\texttt{MPool}}
\newcommand{\ours}{\texttt{MPool}$_S$}
\newcommand{\ourc}{\texttt{MPool}$_C$}
\newcommand{\ourcm}{\texttt{MPool}$_{cmb}$}
\begin{abstract}
Graph Neural networks (GNNs) have recently become a powerful technique for many graph-related tasks including graph classification. Current GNN models apply different graph pooling methods that reduce the number of nodes and edges to learn the higher-order structure of the graph in a hierarchical way. All these methods primarily rely on the one-hop neighborhood. However, they do not consider the higher-order structure of the graph. In this work, we propose a multi-channel \underline{M}otif-based Graph \underline{Pool}ing method named (MPool) that captures the higher-order graph structure with motif and also local and global graph structure with a combination of selection and clustering-based pooling operation. As the first channel, we develop node selection-based graph pooling by designing a node ranking model considering the motif adjacency of nodes. As the second channel, we develop cluster-based graph pooling by designing a spectral clustering model using motif adjacency. As the final layer, the result of each channel is aggregated into the final graph representation. We perform extensive experiments on eight benchmark datasets and show that our proposed method shows better accuracy than the baseline methods for graph classification tasks.

\keywords{Graph Neural Network \and Graph Classification \and Pooling \and Motif.}
\end{abstract}

\section{Introduction}\vspace{-2mm}
By inputting a graph with node attributes and message propagation along the edges, while some GNN~\cite{Kipf} models learn the node-level representation for node classification \cite{Kipf,hamilton2017inductive,velivckovic2017graph}, some others learn graph-level representation for graph classification\cite{gao2019graph, pmlr-v97-lee19c, bianchi2020spectral}. Graph classification is the task of predicting graph labels by considering node features and graph structure. Motivated from the pooling layer in CNNs~\cite{krizhevsky2012imagenet}, graph pooling methods have been used to reduce the number of nodes and edges to capture the local and global structural information of the graph in the graph representation. Current pooling methods usually coarsen the graph in a hierarchical way by reducing the size of the graph in multiple steps and the utilize every pooled graph representation for the final graph representation.

There are mainly two types of hierarchical pooling methods for the graph in the literature: clustering-based and selection-based methods. While clustering-based methods merge similar nodes into super nodes using a cluster assignment matrix, selection-based methods calculate a score for every node, which represents their importance, and select the top $k$ nodes based on the score by discarding other nodes from the graph. All these methods primarily rely on Graph Convolution Networks (GCNs) with layer-wise propagation based on the one-hop neighbors to calculate the assignment matrix in the clustering-based method and score in the selection-based method. Despite the success of these models, there are some limitations. Selection based model mainly focuses on preserving the local structure of the node while the clustering-based method basically focuses on the global structure of the graph. Moreover, while selection-based models may lose information by selecting only some portion of the nodes, clustering-based models may include some redundant information including noise and over-smoothing.\\
Further, the current methods fail to incorporate the higher-order structure of the graph in pooling. There are different ways to model higher-order graph structures \cite{benson2021higher} such as hypergraphs, simplicial complexes~\cite{aktas2021identifying}, and motifs~\cite{milo2002network}. Among them, motifs (graphlets) are small, frequent, and connected subgraphs that are mainly used to measure the connectivity patterns of nodes ~\cite{elhesha2016identification} (see Figure~\ref{fig:motif} for a preview). They capture the local topology around the vertices, and their frequency can be used as the global fingerprints of graphs. 
Although motifs have been used for different graph mining tasks, including classification~\cite{lee2019graph}, and community detection~\cite{li2019edmot}, to the best of our knowledge, they have not been used in graph pooling operations. On the other hand, utilizing these structures for pooling provides crucial information about the structure and the function of many complex systems that are represented as graphs~ \cite{paranjape2017motifs, prill2005dynamic}.\\
In this paper, to address these problems, we propose a multi-channel \underline{M}otif-based Graph \underline{Pool}ing method named (\our) that captures the higher-order graph structure with motif and also local and global graph structure with a combination of selection and clustering-based pooling operation. We utilize motifs to model the relation between nodes and use this model for message passing and pooling in GNN. We develop two motif-based graph pooling models (\ours\ and \ourc): selection and clustering based and then combine these models into one (\ours) to learn both local and global graph structure. For the selection-based graph pooling model, we design a node ranking model considering motif-based relations of nodes. Based on the ranks, we select the top $k$ nodes to create the pooled graph for the next layer. For clustering-based graph pooling, we design a motif-based clustering model that learns a differentiable soft assignment based on learned embedding from the convolution layer. We jointly optimize this function by minimizing usual supervised loss and also unsupervised loss as a relaxation of the normalized mincut objective. However, instead of defining mincut objective on the regular adjacency matrix, we define it in the motif adjacency matrix. After learning the assignment matrix, we group the nodes in the same cluster to create a coarsened graph. Both models take motif into consideration hence incorporating higher-order graph structure in graph pooling operation. We further demonstrate detailed experiments on eight benchmark datasets. Our results show that the proposed pooling methods show better accuracy than the current baseline pooling methods for graph classification tasks.
\vspace{-2mm}
\section{Related Work} \label{sec:rel}\vspace{-2mm}
\textbf{Graph Pooling:}
 Recent pooling methods learn graph representation hierarchically and capture the local substructures of graphs. There are two different hierarchical pooling methods in the literature: clustering-based and selection-based pooling. Clustering-based pooling methods \cite{ying2018hierarchical, bianchi2020spectral,ranjan2020asap,bacciu2021k} do the pooling operation by calculating the cluster assignment matrix using node features and graph topology. After calculating the cluster assignment matrix, they build the coarse graph by grouping the nodes on the same cluster. For example, while DiffPool\cite{ying2018hierarchical} calculates the cluster assignment matrix using a graph neural network, MinCutPool\cite{bianchi2020spectral} calculates the cluster assignment matrix using a multi-label perception.

Selection-based pooling methods \cite{gao2019graph, pmlr-v97-lee19c, gao2021ipool,zhang2019hierarchical,sun2021sugar } compute the importance scores of nodes and select top $k$ nodes based on their scores and drop other nodes from the graph to create the pooled graph. For example, while gPool~\cite{gao2019graph} calculates the score using node feature and a learnable vector, SAGPool \cite{pmlr-v97-lee19c} uses an attention mechanism to calculate the scores. SUGAR \cite{sun2021sugar} uses a subgraph neural network to calculate the score and select top-$K$ subgraph for pooling operation.
All these methods use the classical graph adjacency matrix to propagate information and calculate the score.

\textbf{Motifs in Graph Neural Network}
Motifs are the most common higher-order graph structure used in various graph mining problems. A few works have used motif structure in GNNs as well~\cite{yang2018node, Motif-Attention, peng2020motif, monti2018motifnet}. MCN~\cite{Motif-Attention} creates motif attention mechanism for graph convolution layer to learn node representation. All these methods employ motifs for learning node or subgraph representation. In our proposed method, we use motifs on the graph classification task.
\vspace{-2mm}
\section{Methodology}\label{sec:method} \vspace{-2mm}
In this section, first, we discuss the problem formulation of graph classification and preliminaries. Then we present our motif-based pooling models. \vspace{-3mm} 
\subsection{Preliminaries and Problem Formulation}\label{prbf} \vspace{-2mm}
We denote a graph as $G(V,A,X)$ where $V$ is the node-set, $A\in{\mathbb{R^{N\times N}}}$ is the adjacency matrix, and $X\in \mathbb{R}^{N\times d}$ is the feature matrix with $d$ dimensional node feature and $N$ is the number of nodes in the graph. We denote a graph collection as $(\mathcal{G},Y)$ where $\mathcal{G}=\{G_{0},G_{1}, ..., G_{n}\}$ with $G_i$'s are graphs and $Y$ is the set of the graph labels. In this paper, we work on the graph classification problem, whose goal is to learn a function $f: \mathcal{G} \rightarrow {Y}$ to predict the graph labels with a graph neural network in an end-to-end way.
\\
\textbf{Graph Neural Network for Graph Classification:} GNN for graph classification has two modules: message-passing and pooling. For message-passing operation,  Graph convolution network (GCN) \cite{Kipf} is the most widely used model. GCN is a multilayer neural network that combines the features of each node from its neighbors with propagating the information through the edges as follows:
\begin{equation}
H^{(l+1)}=\sigma{(\tilde{D}^{-\frac{1}{2}}\tilde{A}\tilde{D}^{-\frac{1}{2}}H^{(l)}\theta^{(l)})}\vspace{-1mm}
\end{equation}
where $H^{(l+1)}$  is the node representation matrix for layer $(l+1)$, $\sigma$ is an activation function, $\tilde{A}=A+I$ is the adjacency matrix with self-loop, $\tilde{D}\in{\mathbb{R^{N\times N}}}$ is the normalized degree matrix of $\tilde{A}$, $\theta^{(l)}$ is trainable weight for $l^{(th)}$ layer and $H^{(l)}$ is the input node representation matrix for ${l}^{th}$ layer obtained from previous layer. $H_{0}=X$ is the initial input node feature matrix of the input graph. We utilize GCN for message-passing operation in our model.  

The second module of GNNs for graph classification is the pooling operation that helps to learn the graph features. The main idea behind graph pooling is to coarsen the graph by reducing the number of nodes and edges to encode the information of the whole graph. In the literature, there are two types of hierarchical graph pooling methods: selection-based and clustering-based methods.
Selection-based methods calculate a score (attention) using a scoring function for every node that represents their importance. Based on the calculated scores, the top $k$ nodes are selected to construct a pooled graph. They use a classical graph adjacency matrix to propagate information and calculate the score.

Clustering-based pooling methods learn a cluster assignment matrix $S\in R^{N\times K}$ using graph structure and/or node features. Then, they reduce the number of nodes by grouping them into super nodes by $S\in R^{N\times K}$ to construct the pooled graph at $(l+1)^{th}$ layer as follows\vspace{-1mm}
\begin{equation}\begin{aligned}\label{equ:spool}
A^{(l+1)}=S^{(l)^T}A^{(l)}S^{(l)}, \quad \quad
H^{(l+1)}= S^{(l)^T} H^{(l)}.\end{aligned}\vspace{-1mm}
\end{equation} 

\begin{figure}\vspace{-2mm}
    \centering
    \includegraphics[width=0.8\textwidth]{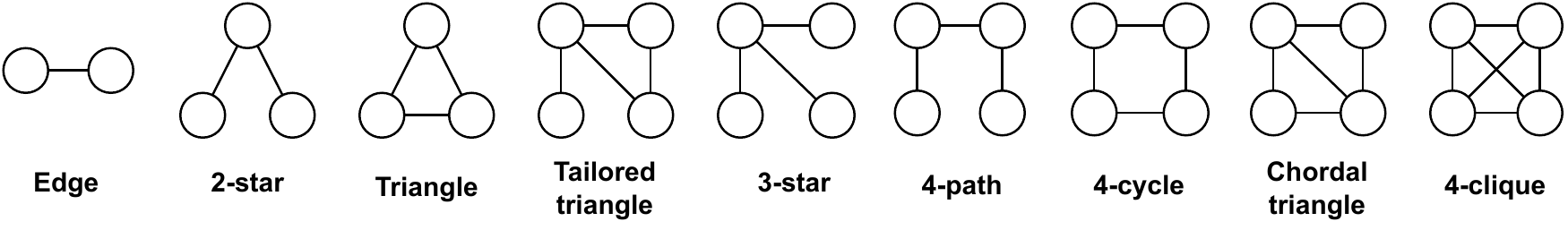}
    \caption{Motif Networks with size 2-4.}
    \label{fig:motif}\vspace{-10mm}
\end{figure}

\textbf{Motifs and Motif-based Adjacency Matrix}:
Motifs (graphlets) are small, frequent, and connected subgraphs that are mainly used to measure the connectivity patterns of nodes ~\cite{ elhesha2016identification}. Motifs of
sizes 2-4 are shown in Figure \ref{fig:motif}. To include higher-order structural information between nodes, we create the motif adjacency matrix $M_{t}$ for a motif $t$ where  $(M_{t})_{i,j}$ represents the \# of the motif containing nodes $i$ and $j$. 
\vspace{-2mm}
\subsection{Motif Based Graph Pooling Models} 
\begin{figure}[t]
    \centering{
    \includegraphics[width=.99\textwidth]{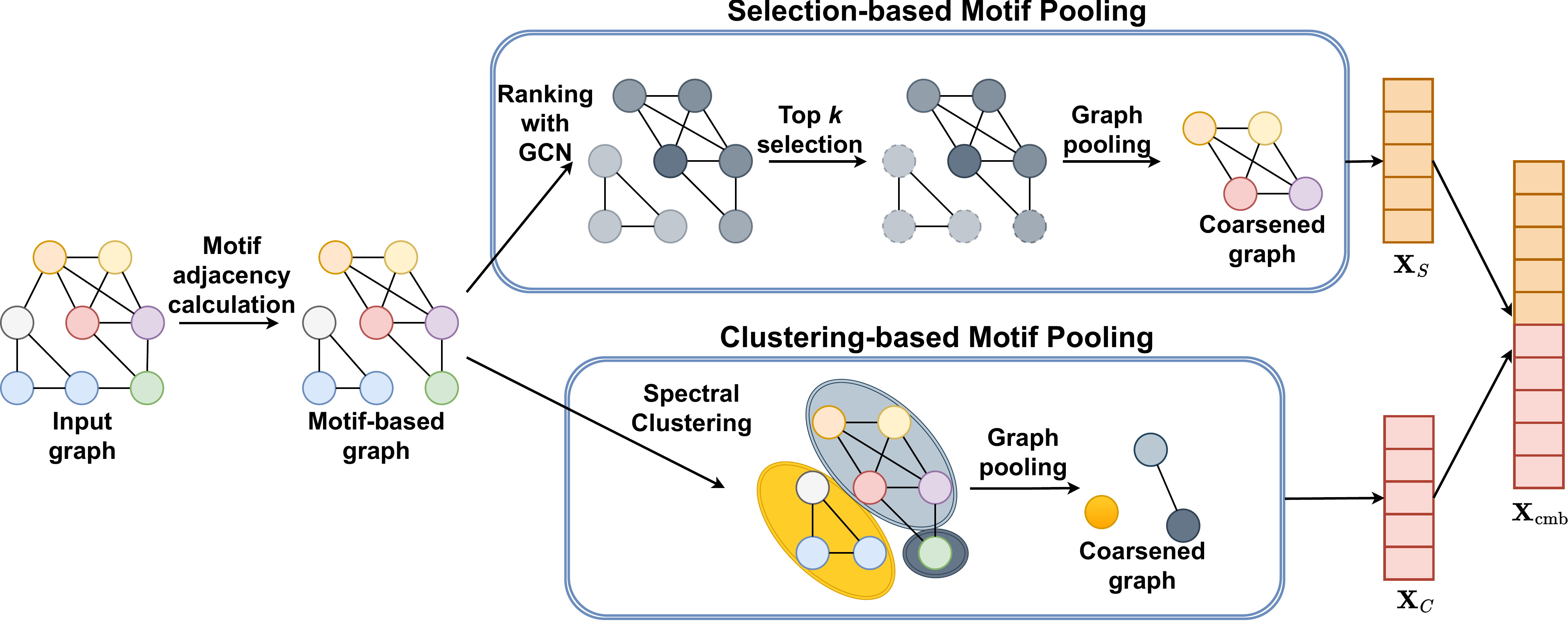}}
    \caption{An illustration of our motif-based pooling methods. }
    \label{fig:select}\vspace{-6mm}
\end{figure}

\vspace{-1mm}
We propose a hierarchical pooling method based on motif structure. As the first layer, graph convolution (GCN) takes the adjacency matrix $A$ and feature matrix $X$ of the graph as input and then updates the feature matrix by propagating the features through the neighbors and aggregating features coming from adjacent nodes. After getting the updated feature matrix from the convolution layer, our proposed graph pooling layer, \our, operates coarsen on the graph. These steps are repeated $l$ steps, and outputs of each pooling layer are aggregated with readout function~\cite{cangea2018towards} to obtain a fixed-sized graph representation. After concatenating the results of readouts, it is fed to the multi-layer perceptron (MLP) layer for the graph classification task. We propose Three types of motif-based graph pooling methods: (1) \ours\ is the selection-based method, (2) \ourc\ is the clustering-based method, and (3) \ourcm\ is the combined model. These are illustrated in Figure~\ref{fig:select}.

In this paper, we adopt the model architectures from SAGPool~\cite{pmlr-v97-lee19c} as the selected based and MinCutPool~\cite{bianchi2020spectral} as the clustering-based model. On the other hand, our method is compatible with any graph neural network that we show later in our experiment section. 
\\
\textbf{A. Selection-based Pooling via Motifs: \ours}\label{nsbm}
Previous selection-based methods~\cite{gao2019graph, pmlr-v97-lee19c} do the pooling operation using a classical adjacency matrix. However, higher-order structures like motifs show great performance on graph convolution network~\cite{lee2019graph} and are also important structures for graph classification. Therefore, while calculating attention scores of nodes, considering motif-induced neighborhoods to integrate information could provide more relevant information about graph structure~\cite{lee2019graph}. In our selection method, we first calculate the motif adjacency matrix for a certain motif type, e.g., triangle, from the original graph as we discuss in Section~\ref{prbf}. Then, we apply the graph convolution network to the motif adjacency matrix and calculate the motif attention score for each node using GNN. Based on these scores, we select the top $k$ nodes for pooling and construct the coarsened graph using the pooling function. Figure~\ref{fig:select} presents the overview of our selection-based graph pooling method. 
\\
\textbf{Motif attention}\label{mas}  
We calculate the motif attention score of nodes to select nodes to drop and nodes to retain. We use graph convolution to calculate the attention where we use node attributes and also motif-based graph topological information instead of pair-wise edge information. 
 \textit{motif attention} score is defined as follows
\begin{equation}\vspace{-1mm}
   \begin{aligned}
    X=GNN(X,\tilde{A};\theta_{GNN}) \\
    Z=\sigma{(D^{-\frac{1}{2}}\tilde{M}D'^{-\frac{1}{2}}X\theta_{att})}
  \end{aligned}\vspace{-1mm}
\end{equation}
where $\tilde{A}$ is the normalized adjacency matrix and $\theta_{GNN}$ is the learnable parameter, $\sigma$ is an activation function, $M'\in{\mathbb{R^{N\times N}}}$ is the motif adjacency matrix with self loop where $\tilde{M}= M+I_N$, $D'\in{\mathbb{R^{N\times N}}}$ is the degree matrix of $M$, and $\theta_{att}\in{R^{d\times 1}}$ is the parameter matrix for pooling layer. Since we use graph features and motif adjacency matrix with convolution for motif attention score, the output of pooling is based on higher-order graph structures and features.
\\
\textbf{Pooling:}\label{map} 
Based on the motif attention score, we select the top $k$ nodes from the graph following the node selection method in~\cite{gao2019graph}. The top $k=\alpha \times{N}$ nodes are selected based on the $Z$ value where $\alpha$ is the pooling ratio between 0 and 1. Thus, we obtain the pooling graph as follows
\begin{equation}\vspace{-1mm}
    \begin{aligned}
     idx=topK(Z,[\alpha \times N])\\
     X_{out} = X_{idx,:,}\odot Z_{idx},  \quad A_{out}=A_{idx,idx}
    \end{aligned}\vspace{-1mm}
\end{equation}
where $idx$ is the indices of the top $k$ nodes from the input graph which is returned by $topK$ function, $X_{idx}$ is the features of the selected $k$ nodes, $Z_{idx}$ is the motif attention value for those nodes. ${\odot}$ is the element-wise broadcasted product, $A_{idx,idx}$ is row and column wised indexed matrix, $A_{out}$ is the adjacency matrix
and $X_{out}$ is the new feature matrix of the pooled graph.
\\\textbf{B. Clustering-based Pooling via Motifs: \ourc}\label{cluster} 
In this paper, as the base for our clustering-based pooling methods, we use MinCutPool \cite{bianchi2020spectral} that is defined based on Spectral clustering (SC) by minimizing the overall intra-cluster edge weights. MinCUTpool proposes to use GNN with a custom loss function to compute cluster assignment with relaxing the normalized minCut problem.  However, they consider only the regular edge-based adjacency matrix to find clusters. On the other hand, considering edge-type relations between nodes may result in ignoring the higher-order relations. Including higher-order relations like motifs for clustering may produce better groups for pooling. 

In our clustering-based method, we calculate the cluster assignment matrix $S$ utilizing motif adjacency information. We adopt spectral clustering method~\cite{bianchi2020spectral} where we use multi-layer perceptron (MLP) by inputting node feature matrix $X$. We use the softmax function on the output layer of MLP. This function maps each node feature $X_{i}$ into the $i^{th}$ row of a soft cluster assignment matrix $S$
\begin{equation}\label{equ:6}
    S  =MLP(X;\theta_{MLP}) = softmax(ReLU(XW_1)W_2)
\end{equation}

However, as it is seen in Equation~\ref{equ:6}, we do not use adjacency but use attributes of nodes obtained from the convolution part. Therefore, to include motif information in the pooling layer, we use the motif adjacency matrix in the convolution layer while passing the message to neighbors as $X=GNN(X,\tilde{M};\theta_{GNN})$ where $\tilde{M}$ is the normalized motif adjacency matrix which we mention in ~\ref{mas}, and $\theta_{GNN}$ and $\theta_{MLP}$ are learnable parameter. 

We also incorporate motif information in the optimization. Parameters of the convolution layer and pooling layer are optimized by minimizing a loss function $\mathcal{L}$ including the usual supervised loss function $\mathcal{L}_s$ and also an unsupervised loss function $\mathcal{L}_u$ as  $\mathcal{L}_u=\mathcal{L}_{c}+\mathcal{L}_{o}$ 
where \begin{equation}\label{equ:7}\mathcal{L}_{c}=-\frac{Tr(S^{T}MS)}{Tr(S^{T}DS)} \quad \textrm{and} \quad \mathcal{L}_o=\left\Vert \frac{S^{T}S} { || S^{T}S||_F }  -\frac{I_K}{\sqrt{K}} \right\Vert_{F}\end{equation} $\mathcal{L}_{c}$ is the cut loss that encourages strongly connected nodes in motif adjacency to be clustered together.
${L}_{o}$ is the orthogonality loss, which helps the clusters to become similar in size. $I_K$ is a (rescaled) clustering matrix $I_{K} = \widehat{S}^T \widehat{S}$, where $\widehat{S}$ assigns exactly $N/K$ points to each cluster.  
After calculating the cluster assignment matrix, we compute the coarsened graph adjacency matrix and attribute matrix using Equation~\ref{equ:spool}.
\\ \textbf{C. Combined model: \ourcm}
Selection-based models mainly focus on preserving the local structure of the node by selecting top-$K$ representative nodes while cluster-based methods basically focus on the global structure of the graph by assigning nodes into $K$-clusters. To utilize the benefits of the selection-based and cluster-based models at the same time, we combine our selection-based and cluster-based motif pooling model into one model. As a result graph representation from the combined model encoded local structure information from the selection-based model and the global structure model from the cluster-based model. In this model we concatenate the graph-level representation from the selection-based motif pooling method and cluster-based motif pooling method into one final representation as follows:
\begin{equation}
    X_{cmb}=X_{S} \smalloplus X_{C}
\end{equation}where $X_{S}$ is the graph-level representation from \ours\ model and $X_{C}$ from \ourc\ method and, $\smalloplus$ is the concatenation operation.
\vspace{-2mm}
\subsection{Readout Function and Output Layer}\vspace{-2mm}
To get a fixed-sized representation from different layers' pooled graph, we apply a readout function\cite{pmlr-v97-lee19c} that aggregates the node features as follows: $Z=\frac{1}{N}\sum_{i=1}^{N} x_{i}|| \overset{N}{\underset{i=1}{\max}}\ x_{i}$
where $N$ is the number of nodes,  $x_{i}$ is the $i^{th}$ node feature and $||$ denotes concatenation. After concatenating the results of all readout functions as a representation of the graph, it is given as an input to a multilayer perceptron with the softmax function to get the predicted label of the graph as $\hat{Y}=softmax(MLP(Z))$ where $Z$ is the graph representation.  For graph classification, parameters of GNNs and pooling layers are optimized by a supervised loss as
$
\mathcal{L}_{s}= -\sum_{i=1}^{L} \sum_{j=1}^{C}Y_{i,j} log \hat{Y}_{i,j}
$
where $Y$ is the actual label of the graph.

\vspace{-2mm}
\section{Experiment}\label{sec:exp} \vspace{-2mm}
We evaluate the performance of our models in graph classification problems and compare our results with the baseline methods for selection-based and clustering-based on different datasets. We also give the results for the variation of our model by utilizing different message-passing models. Further, we analyze the effect of the motif types on the results of the pooling. More experiments can be found on supplements. 
\\
\textbf{Dataset}: We use eight benchmark graph datasets in our experiments commonly used for graph classification~\cite{morris2020tudataset}. Among these, three datasets are social networks (SN); IMDB-BINARY, REDDIT-BINARY, and COLLAB, and five other datasets are biological and chemical networks (BN);D\&D, PROTEINS (PROT),NCI1, NCI109, and Mutagenicity(MUTAG) .\\
\textbf{Baseline}: We use five graph pooling methods as baseline methods. Among them, gPool \cite{gao2019graph} and SAGPool \cite{pmlr-v97-lee19c} are selection-based method and MinCutPool (MCPool) \cite{bianchi2020spectral}, DiffPool \cite{ying2018hierarchical} and ASAP \cite{ranjan2020asap} are clustering-based method. \\
\textbf{Experimental setup:} To evaluate our models for the graph classification task, we randomly split the data for each dataset into three parts. We use 80\% data for the training set, 10\% data for the validation set, and 10\% data for the test set. We do the splitting process 10 times using 10 random seed values. We implement our model using PyTorch and PyTorch Geometric library. For optimizing the model, we use Adam optimizer~\cite{kingma2014adam}. In our experiments, we take node representation size as 128 for all datasets. Our hyperparameters are as follows: learning rate in \{1e-2, 5e-2, 1e-3, 5e-3, 1e-4, 5e-4\}, weight decay in \{1e-2, 1e-3, 1e-4, 1e-5\}, and pooling ratio in \{1/2, 1/4\}. We find the optimal hyperparameters using grid search. We run the model for a maximum of 100K epochs, and there is an early stopping condition if the validation loss does not improve for 50 epochs. Our model architecture consists of three blocks, and each block contains one graph convolution layer and one graph pooling layer like \cite{pmlr-v97-lee19c}. We use the same model architecture and hyperparameters with MinCuT and SAGPool models.
\vspace{-2mm}
\subsection{Overall Evaluation }\label{classification} \vspace{-2mm}
\textbf{Performance on Graph Classification:}
In this part, we evaluate our proposed graph pooling methods for the graph classification task on the given eight datasets. Each dataset contains a certain number of input graphs and their corresponding label. In the graph classification task, we classify the input graph by predicting the label of the graph. We use node features of the graph as the initial features of the model. If a dataset does not contain any node feature, we use node degrees as initial features using one-hot encoding. Table ~\ref{table:accSelect} and Table~\ref{table:accClust} show the average graph classification accuracy, standard deviation, and ranking of our models and other baseline models for all datasets. We can observe from the tables that our motif-based pooling methods consistently outperform other state-of-art models, and our models get the first rank for almost all datasets.

Table~\ref{table:accSelect} shows the results for our motif-based models and other graph pooling models on biochemical datasets. We obtain the reported results for gPool and DiffPool from the SAGPool paper since our model architecture and hyperparameters are the same as SAGPool. Also, for the ASAP method, we obtain the results from the initial publication (``-") means that results are not available for that dataset. As we see from the table, \ourcm\ gives the highest result for all biochemical networks. In particular, \ourcm\ achieves an average accuracy of 81.2\% on D\&D and 77.4\% on NCI1 datasets which are around 4\% improvements over the \ourc\ method as the second-best model. We can also see \ourcm\ gives very good accuracy compared to baseline models for all biochemical datasets. Especially for D\&D, NCI1, and NCI109 datasets \ourcm\ gives 5.8\%, 5.8\%, and 3.9\% improvements over the best model of baseline models for these datasets. From this result, we can say that incorporating global and local structures of the graph in the combined model gives better results for graph classification on biochemical data. We further calculate the average rank for all models, where our model \ourcm\ average rank is the lowest at 1 and our model \ourc\ is the second lowest. 

Table~\ref{table:accClust} shows the performance comparison with our models and other baseline models on social network datasets. As we see from the table, our proposed methods outperforms all the baseline methods for all datasets except ReDDIT-BINARY, where our model is the second best with giving very close to the first one, SAGPool. For IMDB-BINARY and REDDIT-BINARY \ourcm\ model gives better accuracy than the \ours\, and \ourc\ model while for COLLAB dataset \ourc\ give much higher accuracy than our other two models. For both types of datasets our selection-based method \ours\ gives better accuracy than the selection-based baseline methods SAGPool and gPool for most of the datasets. In particular, \ours\ achieves an average accuracy of 77.21\% on D\&D and 76.42\% on Mutagenity datasets which is around 2\%  improvement over the SAGPool method which is our base model. Similarly, our cluster-based model outperforms the baseline methods of cluster-based methods for most of the datasets. Especially, \ourc\ achieves an average accuracy of 83.62\% on COLLAB datasets, which is around 5\%  improvement over the ASAP method as the second-best model and around 14\% improvement over the MinCutPool, which is our base model.

\begin{table}[t]
\centering

\caption{ \centering{ Comparison of our models with baseline  pooling methods for Biochemical Dataset.}}
\label{table:accSelect}

\begin{tabular}{c c c c c c c}
 \hline
 \textbf{Model}  &\textbf{D\&D}&\textbf{NCI1}&  \textbf{NCI109} & \textbf{PROT} & \textbf{Mutag} & \textbf{ Rank}\\ \hline\hline
  gPool &75.0$\pm 0.9$/7 & 67.0$\pm 2.3$/7 & 66.1 $\pm 1.6$/7& 71.1 $\pm 0.9$/7 & 71.9 $\pm 3.7$/8 & 7.2\\
  SAGPool &75.7$\pm 3.7$/6 &68.7$\pm 3.0$/6  & 71.0$\pm 3.4$/4& 72.5 $\pm 4.0$/6 & 74.9$\pm 3.9$/7 & 5.8\\
  MCPool & 76.7$\pm 3.0$/5 & 73.1 $\pm 1.4$/3 & 71.5 $\pm 2.7$/3 & 76.3 $\pm 3.6$/3 & 75.9 $\pm 2.7$/5& 3.8\\
  DiffPool & 66.9 $\pm 2.4$/8 & 62.2 $\pm 1.9$/8 & 62.0$\pm 2.0$/8& 68.2 $\pm 2.0$/8& 77.6 $\pm 2.6$/3 & 7.2\\
  ASAP & 76.9 $\pm 0.7$/4 & 71.5$\pm 0.4$/4 &70.1 $\pm 0.6$/6 & 74.2 $\pm 0.8$/4 &- & 4.5\\\hline \hline
  \ourcm & \textbf{81.2} $\pm 2.1$/1  & \textbf{77.4}$\pm$ 1.9/1 & \textbf{73.5}$\pm 2.5$/1& \textbf{79.3} $\pm 3.3$/1 &\textbf{79.6} $\pm 3.7$/1 & 1\\
  \ours & 77.2 $\pm 4.6$/3 &71.0$\pm 3.4$/5   & 70.8$\pm 2.1$/5 & 72.7 $\pm 4.2$/5& 76.4 $\pm 3.1$/4& 4.4\\
  \ourc & 78.5 $\pm 3.3$/2 &74.4$\pm 1.8$/2 & 73.1$\pm 2.5$/2 & 78.1 $\pm 3.3$/2& 78.8 $\pm 2.1$/2& 2\\\hline\hline

\end{tabular}\vspace{-2mm}
\end{table}

\begin{table}[t]
\centering

\caption{ \centering{ Comparison of our models with baseline  pooling methods for Social Network Dataset.}}
\label{table:accClust}

\begin{tabular}{c c c c  c}
 \hline
 \textbf{Model \quad\quad}  &\textbf{IMDB-B}&\textbf{REDDIT-B}&  \textbf{COLLAB} & \textbf{Avg. Rank}\\ \hline\hline
  gPool &73.40$\pm 3.7$ (3) & 74.70$\pm 4.5$ (6) & 77.58 $\pm 1.6$ (3) & 4\\
  SAGPool &73.00$\pm 4.06$ (4) & \textbf{84.66}$\pm 5.4$ (1)  & 70.1sw$\pm 2.5$ (6) & 3.6 \\
  MinCutPool & 70.78$\pm 4.7$ (7) & 75.67 $\pm 2.7$ (5) & 69.91 $\pm 2.3$ (7) & 6.3 \\
  DiffPool & 68.40 $\pm 6.1$ (8) & 66.65 $\pm 7.7$(7) &74.83 $\pm 2.0$ (4) & 6.3\\
  ASAP & 72.74 $\pm 0.9$ (5) & - &78.95 $\pm 0.7$ (2)  & 3.5\\\hline \hline
  \ourcm & \textbf{74.20} $\pm 2.8$ (1)  & 84.10$\pm$ 5.0 (2) & 74.13$\pm 2.3$ (5) & 2.6\\
  \ours & 73.44 $\pm 3.9$ (2) &83.89$\pm 4.3$ (3)  & 68.95$\pm 2.7$ (8) & 4.3\\
  \ourc & 71.44 $\pm 4.0$ (6) &78.77$\pm 5.0$ (4) & \textbf{83.62}$\pm 5.2$ (1) & 3.6\\\hline\hline

\end{tabular}\vspace{-2mm}
\end{table}

Furthermore, when we compare our selection-based model \ours\ and clustering-based model \ourc\ results from Tables, we can see that \ourc\ outperforms \ours\ for all biochemical datasets. While \ours\ gives better accuracy for two social networks, IMDB-BINARY and REDDIT-BINARY, \ourc\ have 15\% better accuracy than \ours\ on COLLAB dataset. 
\\
\textbf{Ablation Study:} 
While we use GCN as the base model for message passing, our pooling model can integrate other GNN architectures. 
In order to see the effects of different GNN models in our methods, we utilize the other four most widely used convolutional graph models: Graph convolution network (GCN) \cite{Kipf}, Graph-SAGE~\cite{hamilton2017inductive}, GAT~\cite{velivckovic2017graph}, and GraphConv \cite{morris2019weisfeiler}. Table~\ref{table:accGNN} shows average accuracy results for these GNN models using \ours\,\ourc\, and \ourcm\ on NCI1 and IMDB-BINARY datasets. As there is no dense version of Graph attention network(GAT), we use it only for selection-based model \ours\.   For this experiment, we use triangle motifs for the motif adjacency matrix calculation. As we see in the table, the effects of GNN models and which model gives the best result depend on the dataset. For the NCI1 dataset, Graph-SAGE gives the highest accuracy on \ours\ and \ourcm\ model while GraphConv gives the highest accuracy on \ourc model. For IMDB-BINARY, all the graph convolutional models give very close results for all of our pooling models. For \ourc\ and \ourcm\, Graph-SAGE gives better accuracy than the other GNN models while GAT gives the highest accuracy for \ours\ model.

\begin{table}[t]
\footnotesize
\centering
\caption{\ours\ and \ours\ performance with different GNN models.}
\label{table:accGNN}
\begin{tabular}{c|c c| c c| c c}
 \hline 	
& \multicolumn{2}{|c|}{\ours} & \multicolumn{2}{c}{\ourc}& \multicolumn{2}{c}{\ourcm}\\\hline
 \textbf{GNN Model} &\textbf{NCI1  }&\textbf{IMDB-B}  &\textbf{NCI1  }&\textbf{IMDB-B} &\textbf{NCI1  }&\textbf{IMDB-B}\\\hline\hline
 $MPool_{GCN}$ &70.98 & 73.44  &74.44 & 71.44 &76.09 & 73.90 \\\hline
 $MPool_{GraphConv}$ &74.20  &73.50  &75.93  &71.90 & 74.7& 73.00 \\\hline
 $MPool_{SAGE}$ &74.69  & 73.00  &74.13  & 72.22 & 78.80 & 74.00\\\hline
 $MPool_{GAT}$ &67.15 & 74.00 &-&-&- &- \\\hline
 
\end{tabular}\vspace{-2mm}
\end{table}



\begin{table}[t]
\centering
\caption{ \ours\ and \ourc\ performance with different motifs.}
\label{table:accMotif1}
\begin{tabular}{c |c|c c c c }
 \hline 	
Model& motif& DD & NCI1 & Mutagenicity & IMDB-B\\\hline
 \multirow{3}{*}{\ours }&2-star & 77.21 & 69.48 & 70.11  &73.00 \\
 
 & Triangle  &75.63  & 70.98 & 76.42 &73.44  \\
 &2-star+triangle  & 75.63 & 69.82 & 72.39 &69.64 \\\hline
 \multirow{3}{*}{\ourc }&2-star & 78.48 & 73.56 & 73.56 &71.20 \\
 
 & Triangle  &75.80  & 74.44 & 78.77  &71.44  \\
 &2-star+triangle  & 74.21 & 74.20 & 76.00 &70.96 \\\hline
 \multirow{3}{*}{\ourcm }&2-star & 81.20 & 77.36 & 79.60  &74.20 \\
 
 & Triangle  &80.50  & 76.09 & 77.90  &73.90  \\
 &2-star+triangle  & 79.95 & 76.75 & 78.42 &73.40 \\\hline

\end{tabular}\vspace{-2mm}
\end{table}

We further study the effect of the motif type for pooling. In this experiment, we use 2-star, triangle, and a combination of 2-star and triangle motifs, as these motifs are observed the most in real-world networks. We present the graph classification accuracy for different motifs using \ours\,\ourc\, and \ourcm\ in Table \ref{table:accMotif1}.  As we see in the table, we get the highest accuracy for \ours\ and \ourc\ with the triangle motif for three datasets NCI1, Mutagenicity, and IMDB-BINARY. For D\&D, we get the highest accuracy with 2-star motif adjacency on \ours\ and \ourc\. We also observe that for D\&D, the accuracy of the selection-based model does not vary much compared to the clustering-based model. For Mutagenicity, different motifs have a large effect on the accuracy, where triangle motif adjacency gives around 4\% and 3\% higher accuracy than the 2-star motif adjacency for the selection-based method and for the clustering-based model, respectively. For IMDB-BINARY, 2-star and triangle motifs give similar accuracy for both methods, and 2-star+triangle motif adjacency gives less accuracy for the clustering-based method. For our combined model \ourcm\ , the 2-star motif gives the highest accuracy for all datasets whereas other motifs give very close results to the 2-star motif. \\
\vspace{-6mm}
\section{Conclusion} \label{sec:conc} \vspace{-2mm}
In this work, we propose a novel motif-based graph pooling method, MPool, that captures the higher-order graph structures for graph-level representation. We develop graph pooling methods for both types of hierarchical graph pooling models, namely selection-based and clustering-based. We also develop a model where we combined selection-based and cluster-based model. In our selection-based pooling method, we use the motif attention mechanism, whereas, in the clustering-based method, we use motif-based spectral clustering using the mincut loss function.  In both methods, we utilize node feature information and graph structure information together during the learning of the graph feature vector. We show that our proposed methods outperform baseline methods for most of the datasets. 
\vspace{-3mm}

%
%
%
 \bibliographystyle{splncs04}
 \bibliography{mybibliography}

%




\end{document}


\maketitle              
\newcommand{\our}{\texttt{MPool}}
\newcommand{\ours}{\texttt{MPool}$_S$}
\newcommand{\ourc}{\texttt{MPool}$_C$}
\newcommand{\ourcm}{\texttt{MPool}$_{cmb}$}
\subsection{Framework}
\begin{figure*}
   \centering
   \includegraphics[width=0.9\textwidth]{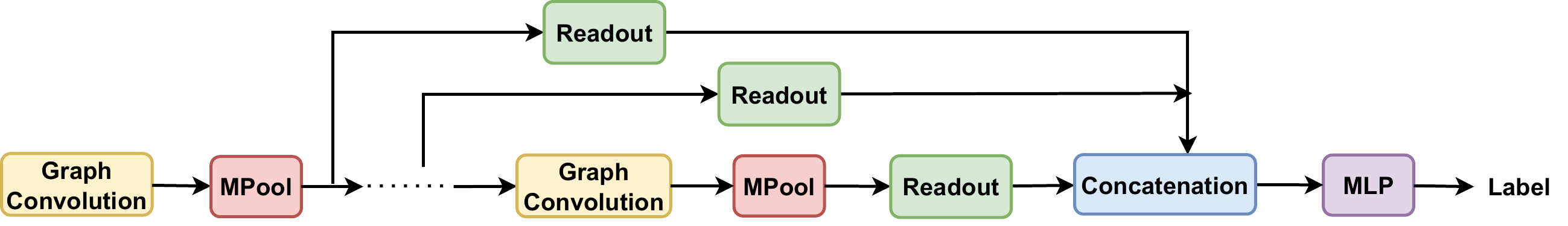}
   \caption{Overall architecture of the proposed GNN model with motif-based graph pooling layer. At each hierarchical layer, we run GCN to update the node embeddings, then give them to the pooling layer as input and generate coarsened graph from pooling layer. Output of each layer is given to readout function and concatenated to get the final embedding of the graph.}
   \label{fig:framework}
\end{figure*}

Figure \ref{fig:framework} shows the overall framework of our proposed model for graph classification. As the first layer, graph convolution (GCN) takes the adjacency matrix $A$ and feature matrix $X$ of the graph as input and then updates the feature matrix by propagating the features through the neighbors and aggregating features coming from adjacent nodes. After getting the updated feature matrix from the convolution layer, our proposed graph pooling layer, \our, operates coarsen on the graph. These steps are repeated $l$ steps, and outputs of each pooling layer are aggregated with readout function~\cite{cangea2018towards} to obtain a fixed-sized graph representation. After concatenating the results of readouts, it is fed to the multi-layer perceptron (MLP) layer for the graph classification task. We propose two types of motif-based graph pooling methods: (1) $MPool_{s}$ is the selection-based method and (2) $MPool_{c}$ is the clustering-based method. These are illustrated in Figure~\ref{fig:select}.

\section{Graph Reconstruction}
In addition to using pooling for the graph classification problem, we further use it in autoencoder (AE) to construct graphs from their representations. We design an AE by optimizing two objectives: (1) node attribute reconstruction and (2) edge reconstruction. For the node attribute reconstruction, we use the mean squared error (MSE) as the optimization function. For the edge reconstruction, the predicted adjacency matrix is scored against the target adjacency matrix using binary cross-entropy (BCE) loss. In our model, we change the last step as one MLP, which is optimized with graph classification task, with two MLPs, one predicts the adjacency matrix with a final sigmoid activation and the other one predicts the node attributes. Both the edge and attribute MLPs use the ReLU activation for the first two layers and are defined as follows
$$\hat{\mathbf{A}}=MLP_{\text{edge}}(V; \theta_{MLP_{\text{edge}}}) \quad \textrm{and } \quad \hat{\mathbf{X}}=MLP_{\text{attribute}}(V; \theta_{MLP_{\text{attribute}}}).$$
\subsection{Complexity Analysis}
The computational complexity of our methods involves two parts that correspond to motif adjacency matrix calculation and pooling layer. Since we use the method from \cite{underwood2020motif} to calculate the motif adjacency matrix, the computation complexity of motif matrix calculation for dense matrices is $O(|V|^{3})$ and for sparse matrices is $O(|V|^{2})$ where $|V|$ represent the number of vertices. In our selection-based method, we next do the graph pooling operation similar to SAGPool \cite{pmlr-v97-lee19c} which has time complexity $O(|V|^{2})$ for dense matrices and $O(|E|)$ for sparse matrices where $|E|$ is the number of edges. Since our implementation is sparse for our selection method, the computational complexity is $O(|V|^{2} + |E|)$. For clustering-based method, we adopt MinCutPool \cite{bianchi2020spectral} that requires $O(K(E + VK))$ computational cost where $K$ is number of cluster in the cluster assignment matrix. It is further implemented by a dense matrix. Hence, the total computational cost for our node cluster-based method is $O(V^{2}+K(E + VK))$.

\section{Datasets}
\begin{table}
\footnotesize
\centering
\caption{Graph statistics. $|G|, V_{avg}$, $E_{avg}$, and $|C|$ denote the number of graph, the average number of nodes and edges, the number of classes in each dataset, respectively.}
\label{table:table1}
\renewcommand{\arraystretch}{1.2}
\begin{tabular}{c c c c c c }
 \hline 	
 &\textbf{Datasets} &\textbf{$|G|$}&\textbf {$V_{avg}$} 
 &\textbf{$E_{avg}$} & 
 \textbf{$|C|$} \\\hline\hline
 &D\&D &1178 & 284.32&715.66&2\\
 \multirow{3}{*}{Biochemical }&NCI1 & 4110 & 29.87 & 32.30&2\\
\multirow{2}{*}{network}& NCI109& 4127 & 29.68 & 32.13&2\\
& PROTEINS& 1113 &39.06 & 72.82&2  \\
 &Mutagenicity & 4337 &30.32 & 30.77&2  \\\hline\hline
 \multirow{2}{*}{Social }& IMDB-BINARY& 1000 & 19.77 & 96.53&2\\
 network&REDDIT-BINARY& 2000 &429.63 & 497.75&2  \\
& COLLAB& 5000 &74.49 & 2457.78&3  \\\hline
\end{tabular}

\end{table}

\section{Performance on Graph Reconstruction:} 
\begin{figure}[t] \vspace{-2mm}
    \centering
    \begin{subfigure}[t]{0.14\textwidth}
        \centering
        \includegraphics[width=0.9\textwidth]{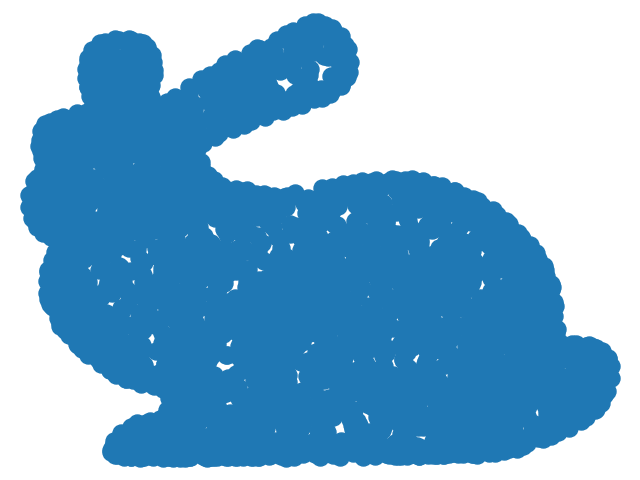}
    \end{subfigure}%
    ~ 
    \begin{subfigure}[t]{0.14\textwidth}
        \centering
        \includegraphics[width=0.9\textwidth]{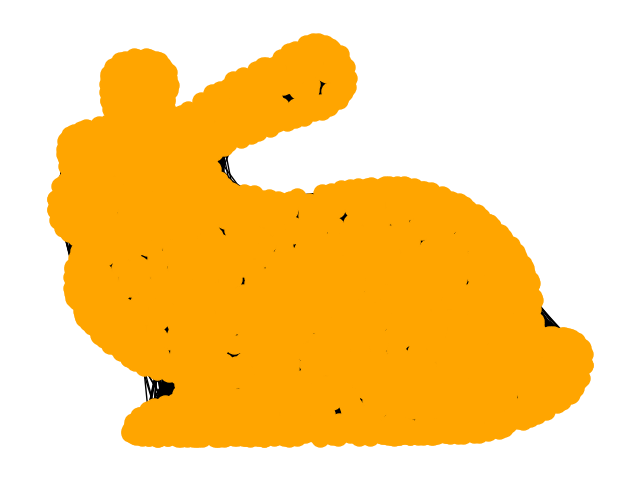}
    \end{subfigure}
    ~
    \begin{subfigure}[t]{0.14\textwidth}
        \centering
        \includegraphics[width=0.9\textwidth]{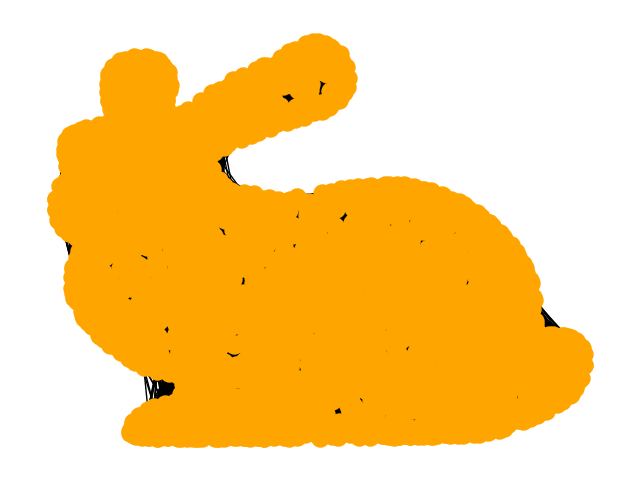}
    \end{subfigure}
    ~ 
    \begin{subfigure}[t]{0.14\textwidth}
        \centering
        \includegraphics[width=0.9\textwidth]{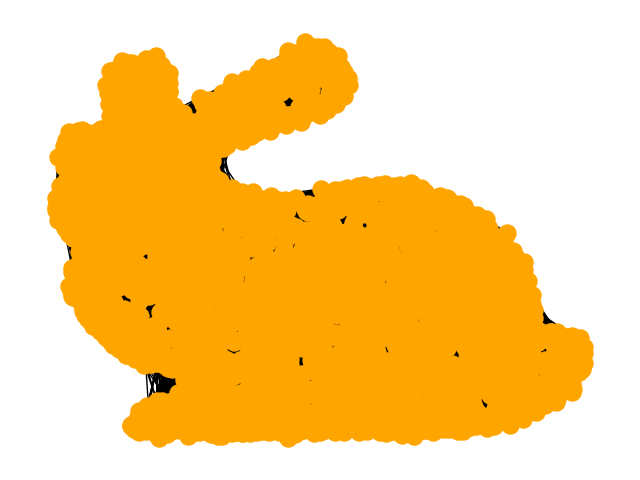}
    \end{subfigure}
     ~ 
    \begin{subfigure}[t]{0.14\textwidth}
        \centering
        \includegraphics[width=0.9\textwidth]{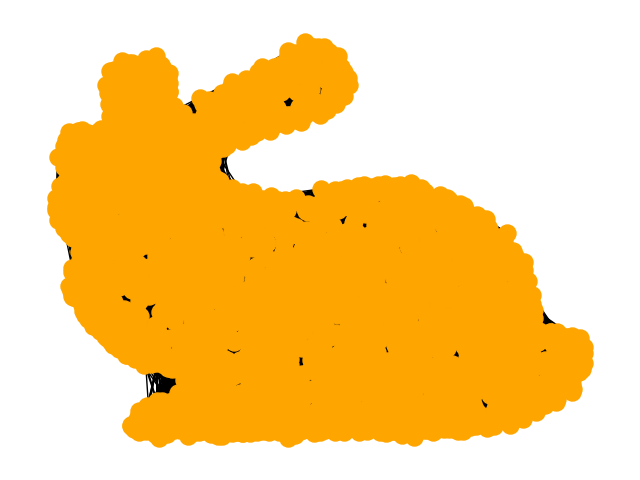}
    \end{subfigure}
     ~ 
    \begin{subfigure}[t]{0.14\textwidth}
        \centering
        \includegraphics[width=0.9\textwidth]{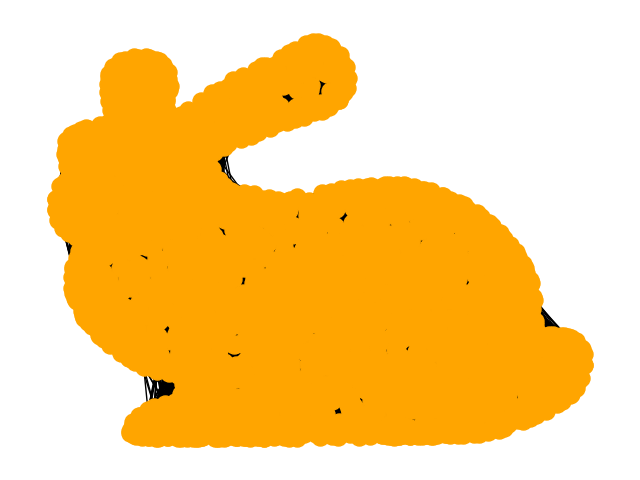}
    \end{subfigure}
    
~
    \begin{subfigure}[t]{0.14\textwidth}
        \centering
        
        \includegraphics[width=0.8\textwidth]{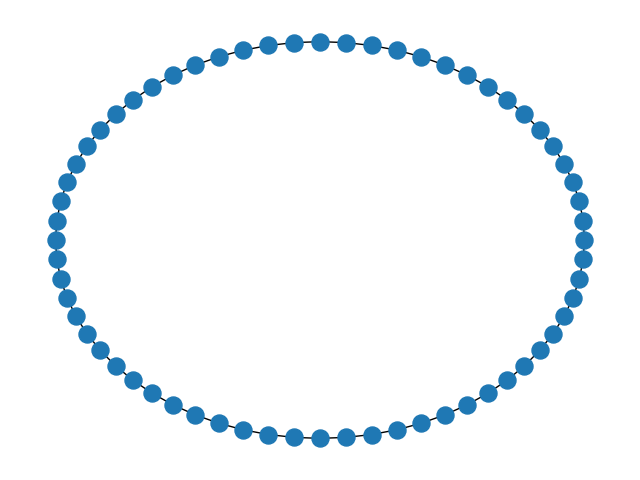}

    \end{subfigure}%
    ~ 
    \begin{subfigure}[t]{0.14\textwidth}
        \centering
        \includegraphics[width=0.8\textwidth]{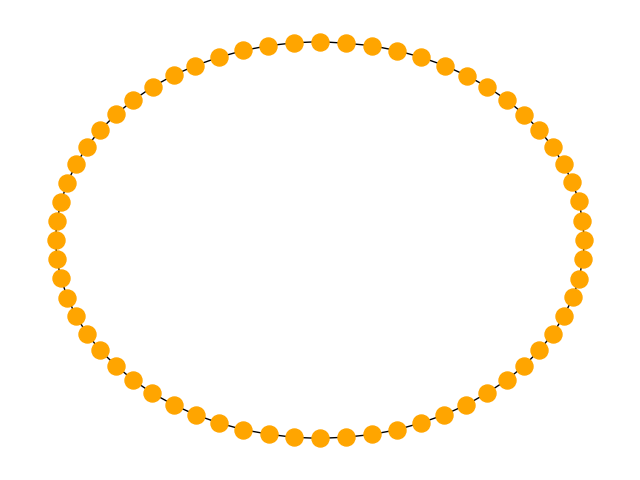}
        
    \end{subfigure}
     ~ 
    \begin{subfigure}[t]{0.14\textwidth}
        \centering
        \includegraphics[width=0.8\textwidth]{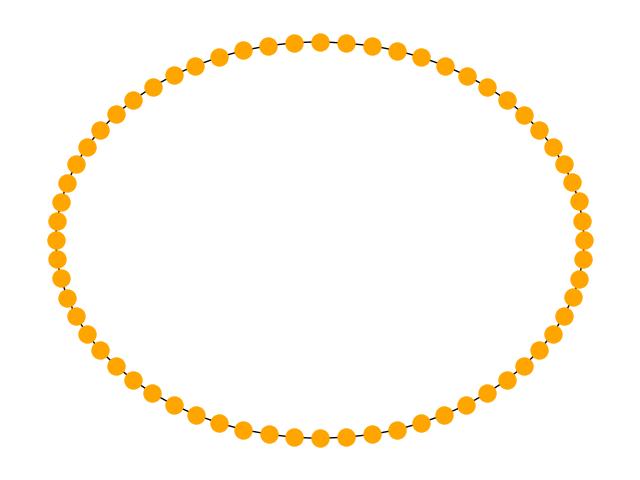}
        
    \end{subfigure}
        ~ 
    \begin{subfigure}[t]{0.14\textwidth}
        \centering
        \includegraphics[width=0.8\textwidth]{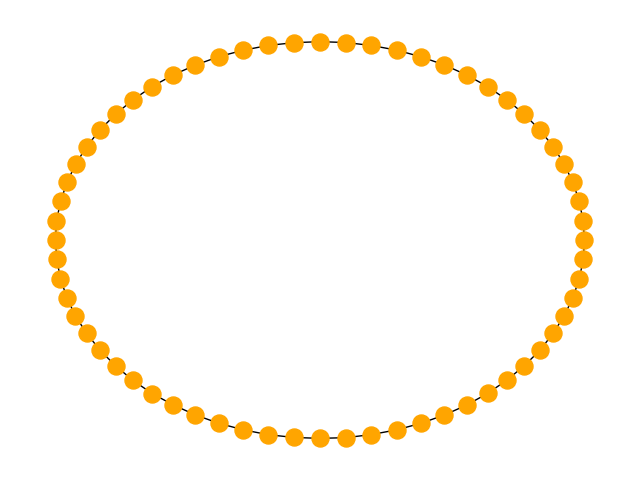}
        
    \end{subfigure}
        ~ 
    \begin{subfigure}[t]{0.14\textwidth}
        \centering
        \includegraphics[width=0.8\textwidth]{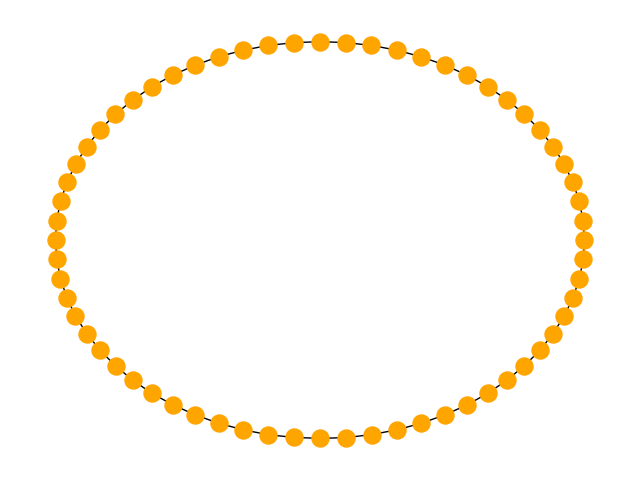}
        
    \end{subfigure}
       ~ 
    \begin{subfigure}[t]{0.14\textwidth}
        \centering
        \includegraphics[width=0.8\textwidth]{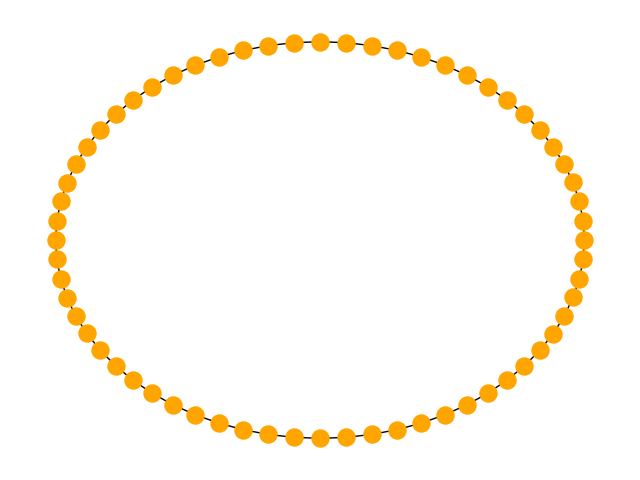}
        
    \end{subfigure}

~
    \begin{subfigure}[t]{0.14\textwidth}
        \centering
        
        \includegraphics[width=0.8\textwidth]{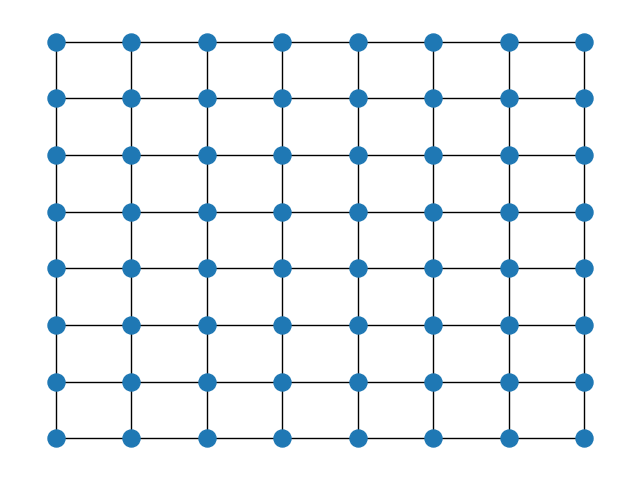}
        
        \caption{Original}
    \end{subfigure}%
    ~ 
    \begin{subfigure}[t]{0.14\textwidth}
        \centering
        \includegraphics[width=0.8\textwidth]{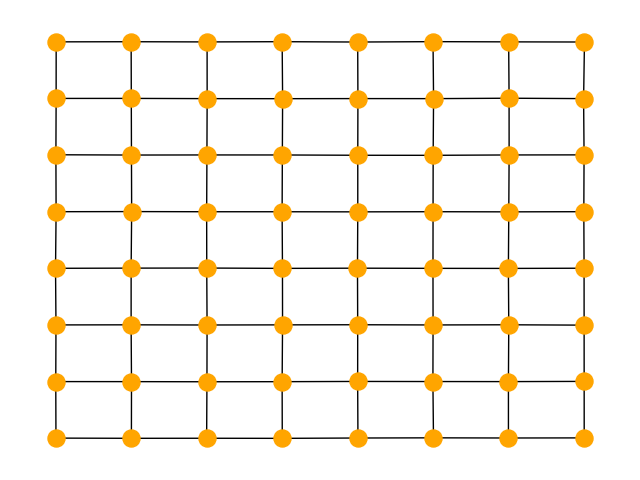}
        \caption{\ourc}
    \end{subfigure}
    ~ 
    \begin{subfigure}[t]{0.14\textwidth}
        \centering
        \includegraphics[width=0.8\textwidth]{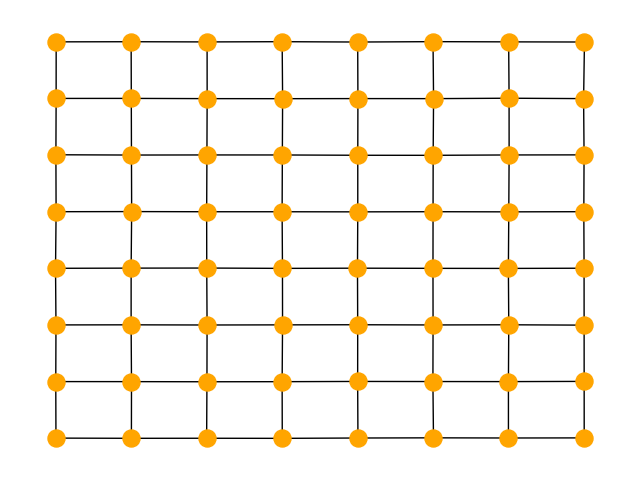}
        \caption{MinCut}
    \end{subfigure}
        ~ 
    \begin{subfigure}[t]{0.14\textwidth}
        \centering
        \includegraphics[width=0.8\textwidth]{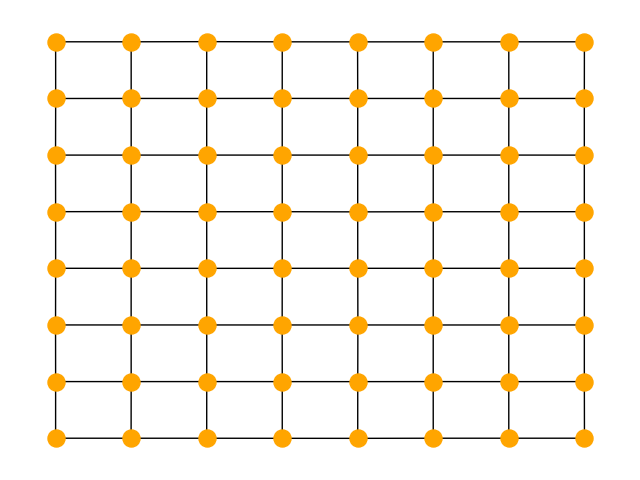}
        \caption{\ours}
    \end{subfigure}
     ~ 
    \begin{subfigure}[t]{0.14\textwidth}
        \centering
        \includegraphics[width=0.8\textwidth]{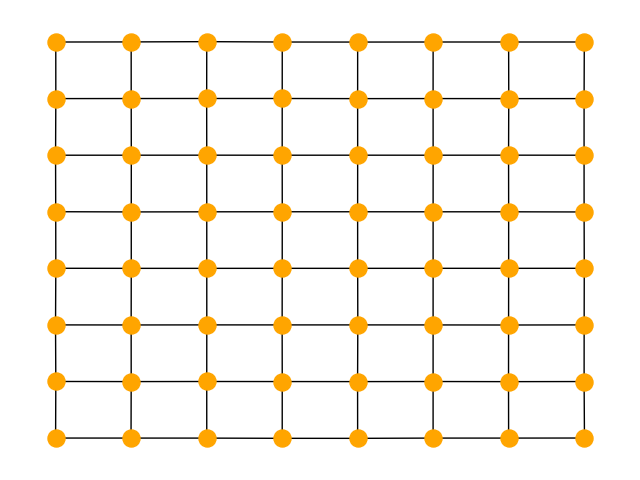}
        \caption{SAGPool}
    \end{subfigure}
    ~ 
    \begin{subfigure}[t]{0.14\textwidth}
        \centering
        \includegraphics[width=0.8\textwidth]{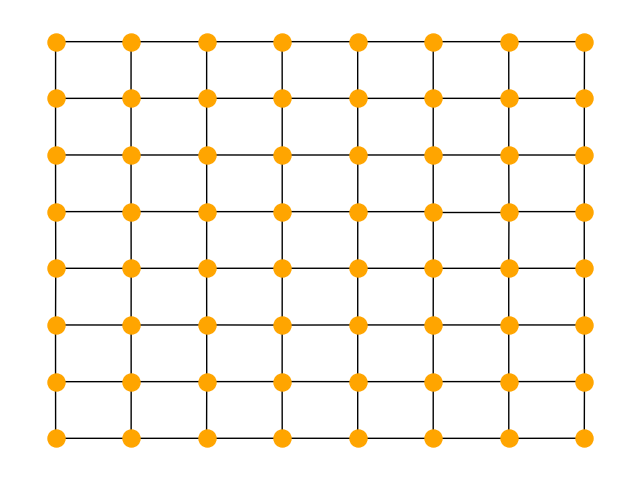}
        \caption{\ourcm}
    \end{subfigure}
    \caption{Graph reconstruction Results after Pooling.}
    \label{fig:graphconstruct}
    \vspace{-5mm}
\end{figure}

We show our graph reconstruction results for our pooling methods and also SagPool and minCUTpool in Figure ~\ref{fig:graphconstruct}. The blue graph is the original, and the orange graph is the reconstructed one. Since the predicted adjacency is probabilistic, if the value is greater than 50\%, we put an edge. All methods have 100\% accuracy for edges on the ring and grid graph and an edge accuracy of 97.96\% on the bunny graph. From these results, we can easily see that our models are able to preserve the content information and graph structure. We also present loss as MSE for node attributes as position in Table~\ref{table:accae}. For the bunny graph, the clustering-based model gives better results. However, the grid graph and ring graph selection-based models have better results, and \ours\ has the lowest node attribute loss for the grid. 

\vspace{-2mm}
\begin{table} \vspace{-2mm}

\centering
\caption{Autoencoder position loss.}
\label{table:accae}
\begin{tabular}{c|c c c c c }
 \hline 	
 Graph &\textbf{$MPool_{S}$}&\textbf{$MPool_{C}$} & \ourcm\ &SAGPool & MinCutPool \\\hline\hline
 Bunny &1.64e-07 & 7.49e-10 & \textbf{2.80e-10}& 1.63e-07 & 7.49e-10 \\\hline
 Ring &4.12e-09  & 9.80e-08  & \textbf{1.62e-09} & 2.44e-09 & 9.79e-08 \\\hline
 Grid &3.24e-09  & 1.11e-07 & \textbf{1.63e-09} & 4.75e-09 & 1.11e-07 \\\hline
 
\end{tabular}
\vspace{-2mm}
\end{table} 

\bibliographystyle{splncs04}
 \bibliography{mybibliography}